%% file: acl_latex.tex
\pdfoutput=1

\documentclass[11pt]{article}

\usepackage[final]{acl}
\usepackage{subcaption}

\usepackage{times}
\usepackage{latexsym}

\usepackage[T1]{fontenc}

\usepackage[utf8]{inputenc}

\usepackage{microtype}

\usepackage{inconsolata}
\usepackage{amsmath}

\usepackage{graphicx}
\usepackage{booktabs}
\usepackage{colortbl} 
\usepackage{tcolorbox}
\usepackage{xcolor}
\usepackage{multirow}
\usepackage{multicol}

\usepackage{setspace}

\usepackage{makecell}

\usepackage{bbding}

\newcommand{\code}[1]{\texttt{#1}}

\title{\textit{VisuoThink}: Empowering LVLM Reasoning with Multimodal Tree Search}

\author{Yikun Wang\thanks{Yikun and Siyin contributed equally}$^1$$^2$,~ Siyin Wang\footnotemark[1]$^1$$^2$, ~Qinyuan Cheng$^1$, ~Zhaoye Fei$^1$, ~Liang Ding$^3$, \\
\textbf{~ Qipeng Guo$^2$$^4$,~ Dacheng Tao$^5$,~ Xipeng Qiu}\thanks{Corresponding Author}$^1$$^2$\\[0.5em]
$^1$ Fudan University $^2$ Shanghai Innovation Institute \\ $^3$ The University of Sydney $^4$ Shanghai AI Laboratory $^5$ Nanyang Technological University\\
\texttt{yikunwang19@fudan.edu.cn}
}

\begin{document}
\maketitle
\begin{abstract}

Recent advancements in Large Vision-Language Models have showcased remarkable capabilities. However, they often falter when confronted with complex reasoning tasks that humans typically address through visual aids and deliberate, step-by-step thinking. While existing methods have explored text-based slow thinking or rudimentary visual assistance, they fall short of capturing the intricate, interleaved nature of human visual-verbal reasoning processes. To overcome these limitations and inspired by the mechanisms of slow thinking in human cognition, we introduce \textbf{\textit{VisuoThink}}, a novel framework that seamlessly integrates visuospatial and linguistic domains. \textit{VisuoThink} facilitates multimodal slow thinking by enabling progressive visual-textual reasoning and incorporates test-time scaling through look-ahead tree search. Extensive experiments demonstrate that \textit{VisuoThink} significantly enhances reasoning capabilities via inference-time scaling, even without fine-tuning, achieving state-of-the-art performance in tasks involving geometry and spatial reasoning. Our code has been open-sourced at \url{https://github.com/ekonwang/VisuoThink}.
\end{abstract}

\section{Introduction}

\input{010-intro-new}

\label{sec:intro}

\section{Related Work}

\input{020-relate}
\label{sec:relate}


\input{030-method}

\label{sec:method}


\input{040-exp}

\section{Discussion}

\input{050-analysis}
\label{sec:analysis}

\section{Conclusion}


We present \textit{\textbf{VisuoThink}}, a multimodal tree search framework enhancing LVLM reasoning through dynamic visual-textual interleaving and predictive rollout search. 
Our approach demonstrates significant improvements across geometry and spatial reasoning tasks without requiring model fine-tuning.  
Empirical results show substantial performance gains on geometry and spatial reasoning benchmarks. 
Our analysis reveals key insights about tool interaction benefits, search space optimization, and supervision strength in multimodal reasoning. 
These findings open new possibilities for advancing LVLM capabilities in complex reasoning tasks.

\section*{Limitations}
Despite its strong performance, \textit{VisuoThink} has several limitations. First, the predictive rollout search process introduces significant computational overhead, making it potentially impractical for real-time applications. Second, our approach particularly relies on tool interactions for stronger capability, which may require more effort in some specific deployment environments. Third, the framework's effectiveness is constrained by the quality of the base VLM's reasoning capabilities - while it enhances performance, it cannot overcome fundamental model limitations. Finally, our evaluation focuses primarily on geometric and spatial reasoning tasks.

\section*{Ethics and Reproducibility Statements}
\paragraph{Ethics} 
We take ethical considerations very seriously and strictly adhere to the ACL Ethics Policy. This paper proposes a test-time slow-thinking framework to improve the multimodal reasoning ability of current LVLMs. All evaluation datasets used in this paper will be publicly available or have been widely adopted by researchers. Thus, we believe that this research will not pose ethical issues.

\paragraph{Reproducibility} In this paper, we discuss the detailed experimental setup, such as hyper-parameters, implementation of algorithm, and statistic descriptions. More importantly, \textit{\textbf{we will open source our code and data in the future}} to help reproduce the experimental results of this paper.

\newpage

\bibliography{custom}

\appendix
\label{sec:appendix}

\input{appendix}

\end{document}

%% file: 010-intro-new.tex
Recent advances in Large Vision-Language Models (LVLMs) \cite{openai2024gpt4ocard, geminiteam2024gemini15} have shown remarkable progress across a variety of tasks. However, these models often struggle with complex reasoning challenges, such as geometric problem-solving \cite{qiao2024we, cherian2024evaluating} or spatial reasoning \cite{ramakrishnan2024doesspatialcognitionemerge, Wu2024MindsEO}, where human problem-solving approaches typically rely on visual aids. For example, when solving geometry problems, humans often iteratively sketch auxiliary lines or visualize intermediate steps, while exploring different reasoning paths - a form of "slow thinking" \cite{kahneman2011thinking} that combines visual and verbal cognitive processes.

\begin{figure}[t!]
    \centering
    \includegraphics[width=0.95\linewidth]{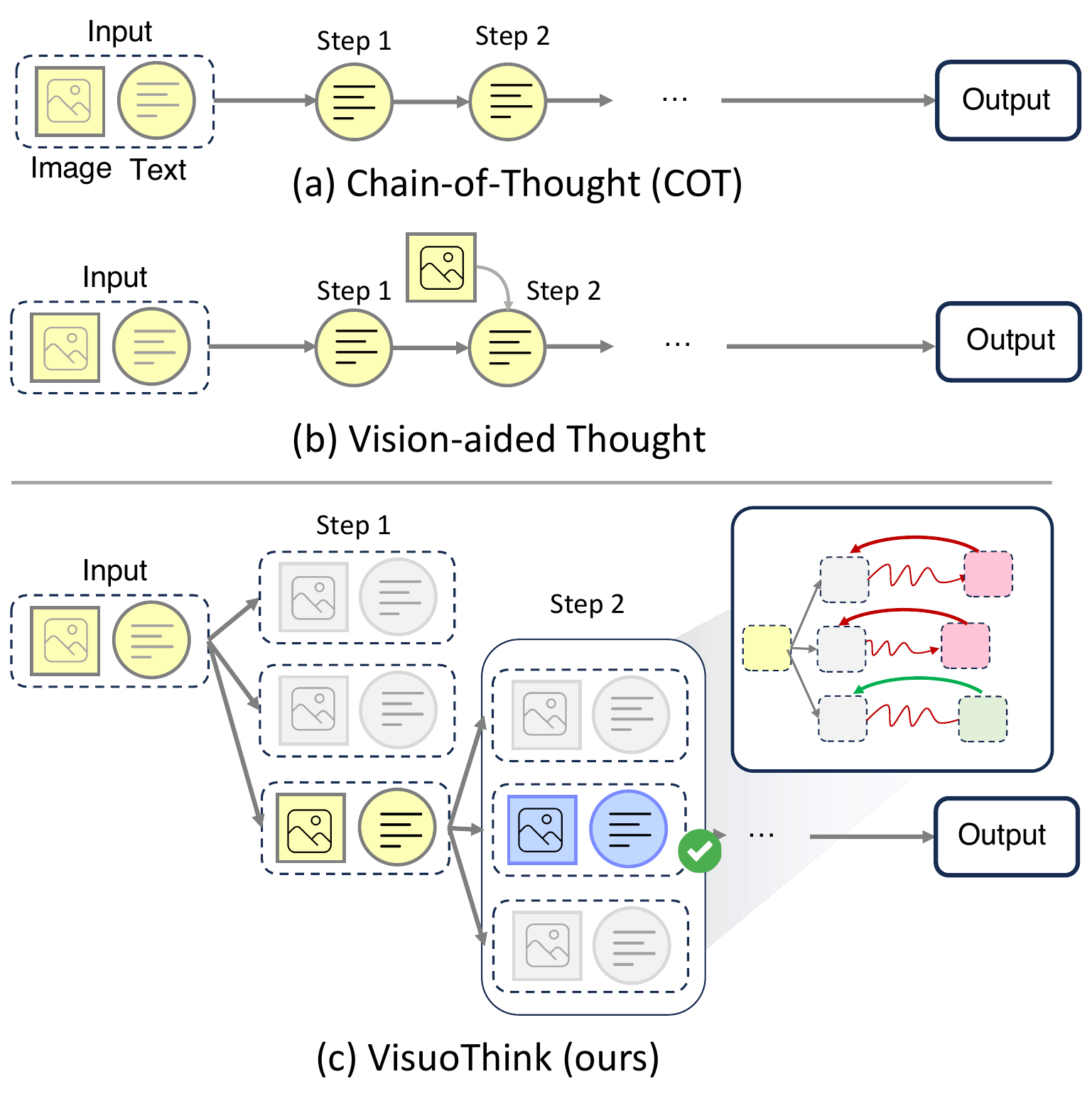}
    \caption{Illustration of Input-Output Prompting, \textit{CoT}, Vision-aided Thought and our \textit{VisuoThink}. Vision-aided Thought often relies on reasoning with one-step or unreliable multi-step visual cues (generated by LVLMs). While \textit{VisuoThink} addresses this gap through tool-augmented visual hints, coupled with a predictive-rollout search mechanism to systematically optimize reasoning capability.}
    \vspace{-0.2in}
    \label{fig:intro}
\end{figure}

With the success of o1 series models \cite{openai2024o1}, researchers have explored language as a medium for implementing slow thinking, coupled with test-time scaling techniques \cite{zeng2024scalingsearchlearningroadmap}. 
Given the inherently multimodal nature of reality, early efforts \cite{Xu2024LLaVACoTLV, Thawakar2025LlamaVo1RS,Yao2024MulberryEM,Du2025VirgoAP} have attempted to extend such deliberative thinking to multimodal reasoning. 
However, even augmented with search strategy, these methods treat visual information merely as static input, relying solely on textual reasoning chains during the reasoning process - creating a "visual blind spot", where the potential for visual information throughout the reasoning process is largely ignored (Fig. \ref{fig:intro}a). 
On the other hand, while approaches like VisualSketchpad \cite{Hu2024VisualSS} and VoT \cite{Wu2024MindsEO} have recognized the importance of visual information by incorporating visual aids in reasoning (Fig. \ref{fig:intro}b), they mainly focus on single-step assistance or simplified visual hints (e.g., emojis).
These methods lack the multi-step visual-textual interleaved reasoning process that characterizes human slow thinking, while failing to explore potential search strategies.

To address these limitations, we propose \textbf{\textit{VisuoThink}}, a multimodal tree search framework that systematically explores multiple reasoning paths with vision-text interleaved thinking at each step.
Unlike previous approaches, Visuothink (Fig. \ref{fig:intro}c) enables multimodal slow thinking through two key innovations: (1) a step-by-step vision-text interleaved reasoning framework that dynamically utilizes multi-step visual aids from tool uses, and (2) a look-ahead tree search algorithm that explores multiple reasoning paths, enabling test-time scaling of the reasoning process. 
Specifically, our look-ahead tree search incorporates a predictive rollout mechanism that simulates the likely outcomes of different reasoning states. This allows the model to prioritize more promising paths and avoid less ones, guiding the reasoning process toward the optimal solution. Through this test-time scaling capability, the model can thoroughly explore and optimize reasoning paths dynamically during inference.

Our empirical evaluation demonstrates that Visuothink significantly outperforms existing methods across various reasoning tasks, particularly in geometry and spatial reasoning domains. 
On Geomeverse, Our methods achieves an accuracy@1 as high as \textit{48.5}\%, with an improvement of as high as \textit{21.8}\% over the state-of-the-art baseline, which particularly shows strong performance of VisuoThink on problems requiring multi-step visual reasoning. 
Through extensive ablation studies, we show that each component of our framework contributes meaningfully to its overall performance.

In summary, our contributions include:

\begin{itemize}
\item We propose a novel reasoning paradigm, multimodal tree search, for multimodal slow thinking that enables dynamic integration of visual and verbal reasoning paths throughout the problem-solving search process.
\item We extend test-time scaling methods to the visual domain by proposing a predictive rollout mechanism that explores and optimizes visual reasoning paths by predicting future states.
\item We demonstrate substantial empirical improvements across multiple reasoning tasks, particularly in geometry and spatial reasoning, with detailed analyses revealing key insights about our approach.
\end{itemize}

%% file: 020-relate.tex
\subsection{Text-centric Reasoning in LVLMs}
With the emergence of o1 models \cite{openai2024o1}, the importance of slow thinking has become increasingly evident \cite{zeng2024scalingsearchlearningroadmap}. 
Several works have attempted to extend this to LVLMs through methods like stage-wise reasoning \cite{Xu2024LLaVACoTLV}, curriculum learning \cite{Thawakar2025LlamaVo1RS}, tree search-based data generation \cite{Yao2024MulberryEM}, and LLM distillation \cite{Du2025VirgoAP}.
However, these methods treat visual information as static input, relying only on textual data during reasoning, which limits their ability to fully leverage multimodal information for complex tasks.

\begin{figure*}[t]
    \centering
    \includegraphics[width=1\linewidth]{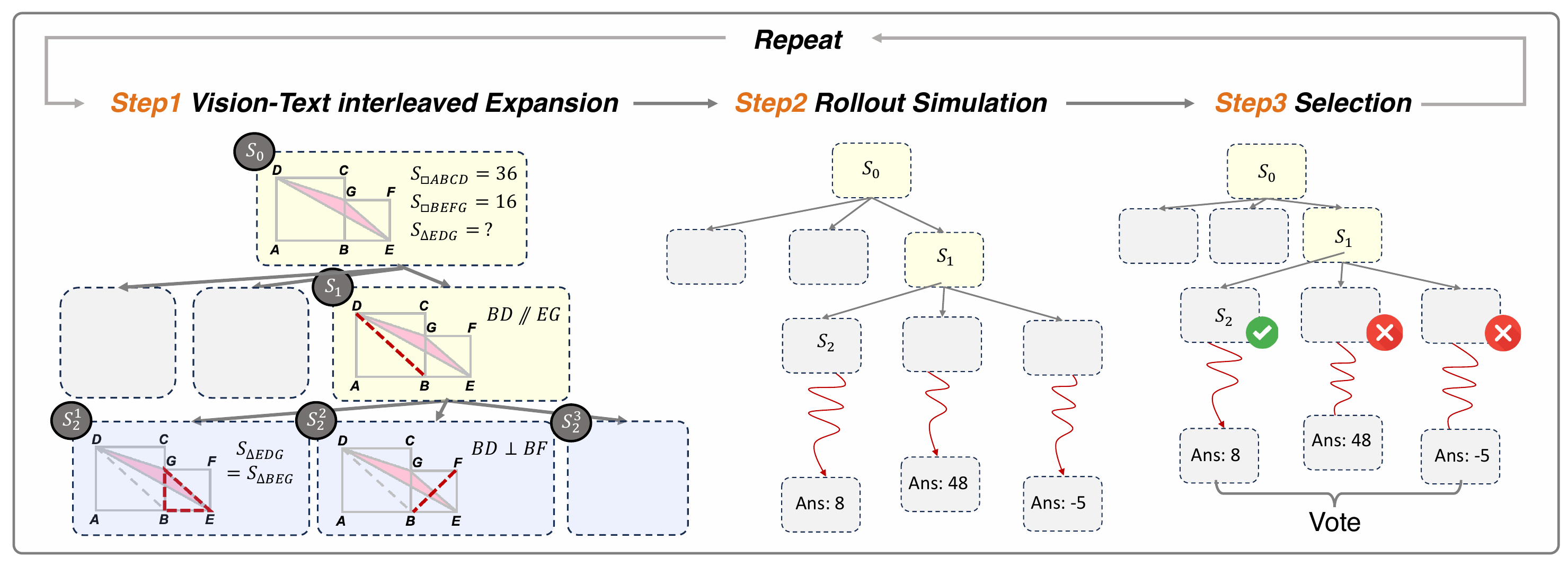}
    \caption{The illustration of our \textit{VisuoThink} framework with three stages: (1) vision-text interleaved expansion: generates candidate paths through vision-text interleaved thinking; (2) rollout simulation: sample candidate reasoning nodes and then perform look-ahead search to better evaluate the value of current states; (3) selection: selects the most promising path via self-voting with results or states from rollout.}
    \label{fig:method}
\end{figure*}

\subsection{Vision-aided Reasoning}

Recent advancements in multimodal reasoning have demonstrated that incorporating visual information provides richer context and hints compared to text-only approaches.
Early studies adopted a two-stage approach, where visual information is first transformed and grounded into text~\cite{Zhang2023MultimodalCR}, graph structures (e.g., scene graphs \cite{Mitra2023CompositionalCP} or knowledge graphs \cite{Mondal2024KAMCoTKA}), or bounding boxes \cite{Lei2024ScaffoldingCT}, followed by reasoning. Other works leverage existing vision models (e.g., segmentation, detection) to process input images into valuable cues for perception, enabling more precise image-understanding with fine-grained visual information \cite{Yang2023MMREACTPC, Zhou2024ImageofThoughtPF, Gao2024CantorIM}.

Another sequence of research focuses on intermediate visual representations to enhance reasoning. For instance, Visual Sketchpad~\cite{Hu2024VisualSS} employs Python-based drawing tools to generate sketches as intermediate visual aids for geometric problems, while VoT~\cite{Wu2024MindsEO} formalizes visual thinking by generating emoji-like textual representations. MVOT~\cite{Li2025ImagineWR} fine-tunes multimodal models to generate images during reasoning, allowing the model to create visual aids dynamically.
Despite these advancements, most existing methods rely on single-step or unreliable visual representations, lacking search mechanisms to test-time scaling through exploring multiple reasoning paths.
In contrast, we develop a multimodal tree search framework that both leverages multi-step visual cues during reasoning and systematically explores reasoning paths through tree search.

\subsection{Test-time Scaling with Tree Search}

Scaling compute at test time has emerged as a powerful strategy to enhance LLMs' reasoning capabilities without increasing model parameters \cite{Snell2024ScalingLT}.
Various approaches including BoN \cite{Gui2024BoNBoNAF, Sun2024FastBD, Amini2024VariationalBA}, guided beam search \cite{Xie2023SelfEvaluationGB, Yu2023OVMOV}, and Monte Carlo Tree Search (MCTS) \cite{Feng2023AlphazerolikeTC,Liu2023DontTA,Chen2024AlphaMathAZ} have been explored for text models, demonstrating improved performance through different search strategies.
However, the exploration of test-time scaling in LVLMs remains limited. Prior work like AtomThink \cite{xiang2024atomthinkslowthinkingframework} has only investigated basic methods such as beam search, with text-only reasoning chains. In contrast, our method introduces vision-text interleaved thinking with look-ahead search, extending test-time scaling to multimodal reasoning.

%% file: 030-method.tex
\section{VisuoThink}


We propose \textit{\textbf{VisuoThink}}, a novel framework for multimodal reasoning that dynamically integrates visual and textual information during the inference process. At its core, our framework implements multimodal slow thinking through a key mechanism: predictive rollout search that allows models to \textit{think} ahead.

\subsection{Vision-Text Interleaved Thinking}

Our framework facilitates vision-text interleaved reasoning through an iterative cycle of \textbf{Thought}, \textbf{Action}, and \textbf{Observation} like existing work \cite{react}, which enables natural and dynamic interactions with external tools.
(1) Thought phase: the model leverages visual information for textual reasoning (such as analyzing patterns based on previously added auxiliary lines) and determines the next step by planning what visual hints should be added to enhance understanding. (2) Action phase: the model executes the planned operations by calling external tools (like using \code{Python} code to draw auxiliary lines or highlight key features) to generate or modify visual information. (3) Observation phase: the model processes the visual feedback from the Action phase, incorporating these new visual hints into  the next reasoning step.

The importance of visual information for LVLM reasoning is highlighted in \textit{VisuoThink},  
which utilize tool invocations to construct \textit{reliable} visual hints step by step in a visual construction process.
This tool-based design allows \textit{VisuoThink} to flexibly adapt to various visual reasoning tasks. Moreover, unlike approaches (e.g. \textit{VisualSketchpad}) that generate all visual aids at once, our step-by-step visual guidance naturally integrates with search techniques, enabling effective test-time scaling.

\subsection{Predictive Rollout Search}
Based on tree search methods and inspired by MCTS, we propose a predictive rollout search mechanism that interleaves visual-text thinking. By anticipating the outcomes of intermediate states, the model can make timely corrections, enabling more accurate and powerful reasoning.
As shown in Figure \ref{fig:method}, at each reasoning step, our framework first generates multiple candidate paths through vision-text interleaved thinking, then simulates these paths to predict their outcomes, and finally selects the most promising path through a self-voting mechanism.

\vspace{-0.05in}

\paragraph{Vision-Text Interleaved Expansion}
In the whole reasoning chain $\textbf{A} = \{\textbf{a}_1, \textbf{a}_2, \dots, \textbf{a}_t\}$, given the current node $\textbf{a}_{t-1}$, the model samples $k$ candidate nodes $\textbf{S}_t = \{\textbf{s}_t^1, \textbf{s}_t^2, ..., \textbf{s}_t^{k}\}$. Each candidate follows the vision-text interleaved thinking process described above, generating a sequence of Thought, Action, and Observation steps. This expansion creates a tree of possible reasoning paths, each representing a different problem-solving strategy.

\vspace{-0.05in}

\paragraph{Rollout Simulation}

Visual reasoning often requires multiple steps to reach a conclusion, making it crucial to evaluate the full potential of each path. For each candidate node $\textbf{s}_t^i$, the model simulates the complete reasoning process to predict final outcomes $\textbf{r}_t^i$, rather than relying solely on immediate state evaluation.
Different from expansion, the simulation extends each candidate node with a single path of vision-text interleaved thinking until reaching a final result.

\vspace{-0.05in}

\paragraph{Selection}
The selection of the optimal path is performed through a self-voting mechanism. The model considers the task description, historical nodes, and the simulated path with predicted results for each candidate node. The selection process can be formalized as:

\begin{small}
\begin{equation}
    \mathbf{Select}(\textbf{S}_t) = \arg\max_{\textbf{s}_t^i \in \textbf{S}_t} \mathbf{Vote}(\textbf{A}_{t-1}, \textbf{s}_t^i, \textbf{r}_t^i)
\end{equation}
\end{small}

where $\textbf{A}_{t-1}$ represents the historical context, $\textbf{s}^i_t$ for the candidate node, and $\textbf{r}^i_t$ is the predicted result or final state. The $\textbf{Select}$ is a heuristic function served by the LVLM model to guide the process. This selection ensures the model pursues the most promising reasoning strategy.

%% file: 040-exp.tex
\section{Solving Geometry with VisuoThink}
\label{sec:geometry}

\begin{table*}[!t]
\centering
\resizebox{.9\textwidth}{!}{
    \begin{tabular}{llccc}
    \toprule
    & Model & \textbf{GPT-4o} & \textbf{Qwen2-VL-72B-Instruct} & \textbf{Claude-3.5-sonnet} \\
    \midrule
    \multirow{5}{*}{\textbf{Geomverse-109}} & CoT & 11.1 & 5.6 & 14.4 \\
    & VisualSketchpad & 8.9 & 6.7 & 16.7 \\
    & VisualSketchpad \textit{+ Equation Solver} & 13.3 & 11.1 & 17.8 \\
    & \cellcolor{gray!20}\textbf{VisuoThink w/o rollout search} (\textit{ours}) 
    & \cellcolor{gray!20}24.4 
    & \cellcolor{gray!20}19.0 
    & \cellcolor{gray!20}26.7 \\
    & \cellcolor{gray!20}\textbf{VisuoThink } (\textit{ours}) 
    & \cellcolor{gray!20}\textbf{28.9} 
    & \cellcolor{gray!20}\textbf{25.6}
    & \cellcolor{gray!20}\textbf{27.8} \\
    
    \midrule
    \multirow{5}{*}{\makecell{\textbf{Geometry3K} \\
    \cite{lu2021intergpsinterpretablegeometryproblem}}} & CoT & 20.8 & 18.8 & 37.5 \\
    & VisualSketchPad & 22.9 & 17.0 & 39.6 \\
    & VisualSketchpad \textit{+ Equation Solver} & 25.0 & 14.9 & 41.7 \\
    & \cellcolor{gray!20}\textbf{VisuoThink w/o rollout search} (\textit{ours}) 
    & \cellcolor{gray!20}27.1 
    & \cellcolor{gray!20}20.8 
    & \cellcolor{gray!20}37.5 \\
    & \cellcolor{gray!20} \textbf{VisuoThink} (\textit{ours}) 
    & \cellcolor{gray!20}\textbf{33.3} 
    & \cellcolor{gray!20}\textbf{25.0} 
    & \cellcolor{gray!20}\textbf{43.8} \\
    \bottomrule
    \end{tabular}
}
\caption{The 1-shot benchmark results (\textit{Accuracy@1}) on Geometry including \textbf{Geomverse-109} and \textbf{Geometry3k} of SOTA large visual language models. For GPT-4o and Claude-3.5-sonnet, we employ newest cutoffs (\textit{gpt-4o-2024-11-20} and \textit{claude-3-5-sonnet-20241022}) separately. The \textcolor{gray!40}{gray} part indicates results from VisuoThink and \textbf{bold} results represent the best performance.}
\label{tab:geometry}
\end{table*}

\begin{table*}[h]
\centering
\resizebox{.95\textwidth}{!}{
    \begin{tabular}{llcccc}
    \toprule
    \multirow{2}{*}{Model}& Dataset & \multicolumn{3}{c}{\textbf{Visual Navigation}} & \textbf{Visual Tiling} \\
    & Subset (Num. Samples) & \textit{level-3 (16)} & \textit{level-4 (31)} & \textit{level-5 (62)} & \textit{level-2 (119)} \\ \midrule
    \multirow{5}{*}{\textbf{GPT-4o}} 
    & CoT & 18.8 & 3.2 & 0.0 & 0.8 \\
    & VoT & 25.0 & 0.0 & 0.0 & 1.7 \\
    & VoT + \textit{Executer} & 62.5 & 9.7 & 4.8 & 12.6 \\
    & \cellcolor{gray!20}\textbf{VisuoThink w/o rollout search} (\textit{ours}) 
    & \cellcolor{gray!20}81.2 
    & \cellcolor{gray!20}32.3 
    & \cellcolor{gray!20}11.3 
    & \cellcolor{gray!20}19.3 \\
    & \cellcolor{gray!20} \textbf{VisuoThink} (\textit{ours}) 
    & \cellcolor{gray!20}\textbf{93.8} 
    & \cellcolor{gray!20}\textbf{61.3} 
    & \cellcolor{gray!20}\textbf{19.4} 
    & \cellcolor{gray!20}\textbf{51.2} \\
    \midrule
    \multirow{5}{*}{\textbf{Qwen2-VL-72B-Instruct}}
    & CoT & 6.7 & 3.2 & - & 0.0 \\
    & VoT & 0.0 & 0.0 & - & 0.8 \\
    & VoT + \textit{Executer} & 25.0 & 3.2 & - & 6.7 \\
    & \cellcolor{gray!20}\textbf{VisuoThink w/o rollout search} (\textit{ours}) 
    & \cellcolor{gray!20}50.0 
    & \cellcolor{gray!20}6.5 
    & \cellcolor{gray!20}- 
    & \cellcolor{gray!20}9.2 \\
    & \cellcolor{gray!20} \textbf{VisuoThink} (\textit{ours}) 
    & \cellcolor{gray!20}\textbf{81.3} 
    & \cellcolor{gray!20}\textbf{12.9} 
    & \cellcolor{gray!20}- 
    & \cellcolor{gray!20}\textbf{20.2} \\ 
    \midrule
    \multirow{5}{*}{\textbf{Claude-3.5-sonnet}}
    & CoT & 37.5 & 3.2 & 0.0 & 0.8 \\
    & VoT & 56.3 & 0.0 & 0.0 & 2.5 \\
    & VoT + \textit{Executer} & 68.8 & 22.6 & 16.1 & 10.1 \\
    & \cellcolor{gray!20} \textbf{VisuoThink w/o rollout search} (\textit{ours}) 
    & \cellcolor{gray!20}81.2 
    & \cellcolor{gray!20}38.7 
    & \cellcolor{gray!20}41.9 
    & \cellcolor{gray!20}80.7 \\
    & \cellcolor{gray!20} \textbf{VisuoThink} (\textit{ours}) 
    & \cellcolor{gray!20}\textbf{93.8} 
    & \cellcolor{gray!20}\textbf{61.3} 
    & \cellcolor{gray!20}\textbf{53.2} 
    & \cellcolor{gray!20}\textbf{84.0} \\
    \bottomrule
    \end{tabular}
}
\caption{The \textit{Pass@1} performance comparison on spatial reasoning benchmarks including \textbf{Visual Navigation} and \textbf{Visual Tiling} across \textit{SOTA} LVLMs. The \textcolor{gray!40}{gray} part indicates results from VisuoThink and \textbf{bold} results represent the best performance. The results of Qwen2-VL-72B-Instruct on Visual Navigation (\textit{k = 5}) are masked out due to its restrained performance on the subset. The results from \textit{VoT} with \textit{Executor} are also reported, where the models utilize the unreliable visual hints generated by themself rather than \textit{executor}, consistent with the \textit{VoT} framework.}
\vspace{-0.1in}
\label{tab:visual}
\end{table*}

The core of our methodology is rooted in multi-step visual information processing and search-based reasoning, enabling LVLMs to address strongly constrained mathematical problems (e.g., geometry challenges) and open-domain scenarios (such as visual navigation and visual tiling in section \ref{sec:spatial_reasoning}).

We formalize geometry problem-solving as a two-phase process integrating \textbf{visual construction} and \textbf{algebraic computation}. In Phase I, the model generates auxiliary lines defined by geometric constraints, such as connecting points $(x_i, y_i)$ and $(x_j, y_j)$, construct a perpendicular or parallel line to form line segments $\textbf{L} = \{l_i\}$. This phase terminates with a \code{AUX-END} token, triggering Phase II, where geometric relationships are translated into solvable equations (e.g., $ax + b = 0$) through \code{Python} code execution.

\vspace{-0.05in}

\paragraph{Task Formulation} LVLM should produce the reasoning trajectory consisting of reasoning steps \( \mathbf{A} = \{\mathbf{a}_t\} \) that leads to the final result \( \textbf{r} \), given the original problem \( \mathbf{Q} \) while taking into account the auxiliary lines \( \mathbf{L} \). The framework operates under a constraint $\sum_{t=1}^{|A|} \| \mathbf{a}_t \| \leq \tau$, where $\mathbf{a}_t$ denotes visual-textual reasoning steps and $\tau$ is the maximum step limit:

\begin{small}
\begin{equation}
   \mathbf{\textbf{A}} \sim \mathcal{P}\left(\{\mathbf{a}_1, \dots, \mathbf{a}_{|A|}, \textbf{r}\} \mid \mathbf{Q}, \mathbf{L}\right) \\
   \text{s.t.}~ \sum_{t=1}^{|\textbf{A}|} \| \mathbf{a}_i \| \leq \tau
\end{equation}
\end{small}

This formulation mirrors human problem-solving by decomposing proofs into executable visual-textual steps, validated via coordinate-based tools like matplotlib and equation solver.

\vspace{-0.05in}

\paragraph{Visual Construction} We emphasize the criticality of incremental visual information for accurate solutions, where multi-step graphical representations originate from the progressive construction of auxiliary lines. This multi-stage approach facilitates search algorithm-enhanced refinement of auxiliary line generation, significantly improving LVLM capabilities in geometric reasoning. Consistent with Sketchpad methodology, we exclusively utilize common \code{Python} libraries (e.g., \textit{matplotlib}) for diagram rendering.

\vspace{-0.05in}

\paragraph{Algebraic Computation} Unlike general tasks, solving geometry problems cannot rely solely on visual construction or the model's inherent capabilities; instead, it necessitates the use of computational tools to achieve precise and accurate results. This requirement stems from the need for exact numerical solutions and the mitigation of potential errors in geometric reasoning. Through systematic integration, like VPD \cite{zhao2023unleashingtexttoimagediffusionmodels}, and VisualStechpad \cite{Hu2024VisualSS}, phase II employs Python code execution for precise computation to mitigate LVLM hallucination risks. Furthermore, the model constructs single-variable algebraic equations based on identified geometric relationships, subsequently invoking equation solvers for numerical resolution.

\subsection{Empirical Results}
\label{subsec:empirical_results_goemetry}

\paragraph{Setup} We conduct comprehensive evaluations on the challenging Geometry3K and Geomverse-109 datasets to demonstrate the methodological superiority. Especially we detail the trajectory of Geomverse-109 dataset synthesis in appendix \ref{appendix:geomverse}. SOTA closed-source models including \textit{gpt-4o-2024-11-20} and \textit{claude-3-5-sonnet-20241022} are leveraged for inference. To ensure architectural diversity, open-source model (e.g., \textit{Qwen2-VL-72B}) were incorporated; however, smaller-parameter open-source variants were excluded due to their capability constraints. And we detail the model and algorithm hyperparameters in appendix \ref{appendix:hyperparameter}.

\begin{figure*}[!t]
    \centering
    \includegraphics[width=1\linewidth]{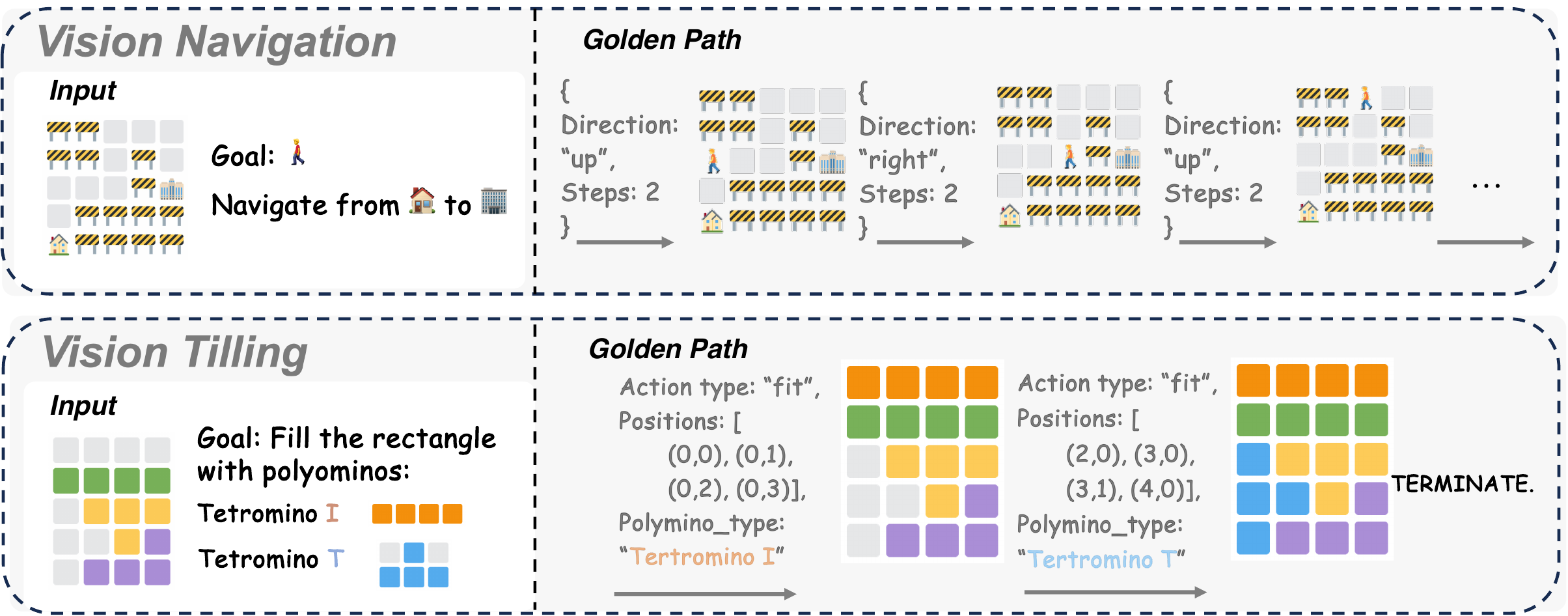}
    \caption{The illustration of spatial reasoning tasks derived from \textit{VoT} \cite{Wu2024MindsEO}, including Visual Navigation and Visual Tiling. LVLM is required to execute a sequence of actions to complete certain goals. Our experimental setting makes them much more challenging and closer to real-environment deployment.}
    \vspace{-0.1in}
    \label{fig:spatial_reasoning}
\end{figure*}

\paragraph{Analysis} Our empirical results reveal that, even without rollout search augmentation, our strategy substantially enhances LVLM reasoning capabilities compared to Chain-of-Thought (CoT) \cite{Mitra2023CompositionalCP} and Visual Sketchpad \cite{Hu2024VisualSS} baselines. Notably, on the Geomverse-109 \cite{geomverse} benchmark, \textbf{\textcolor{black!80}{VisuoThink outperforms CoT and Visual Sketchpad by an average of \textit{17.1}\% and \textit{16.7}\% across all evaluated models, and predictive rollout search further enhances models' performance by an average of 4.1\%}}. Also, the employment of \textit{equation solver} on Visual Sketchpad also increases an average performance of \textit{3.3}\%. This performance gap likely stems from Geomverse's emphasis on geometric relationship construction, where our equation-solving framework help to accurately get intermediate answers and enables efficient resolution of structurally complex problems. The systematic integration of geometric analysis tools further mitigates error propagation inherent in conventional LVLM reasoning baselines. 

\section{Spatial Reasoning with VisuoThink}
\label{sec:spatial_reasoning}

\begin{figure*}[t]
    \centering
    \begin{subfigure}{.45\textwidth}
        \includegraphics[width=\linewidth]{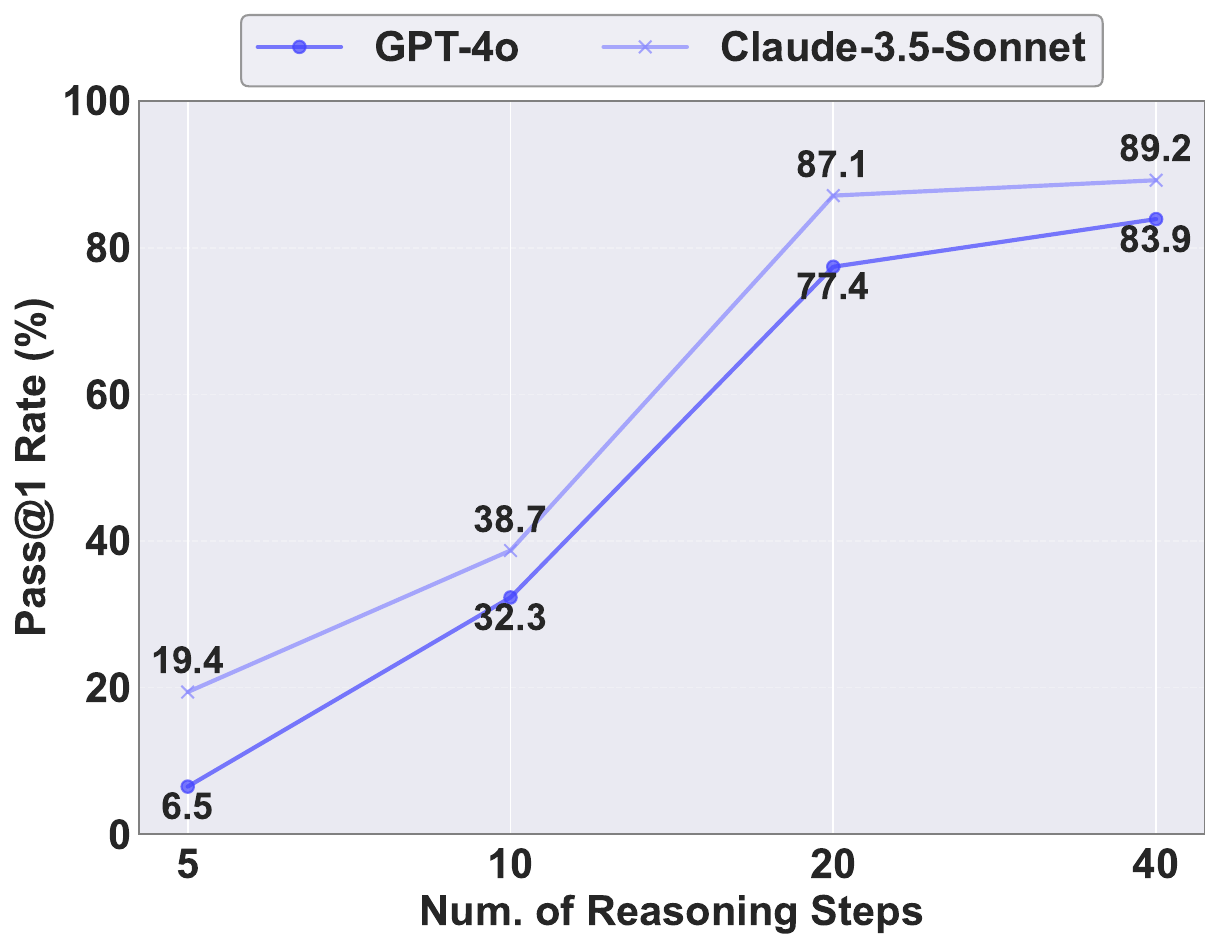}
        \label{figs/api_calls}
    \end{subfigure}
    \begin{subfigure}{.45\textwidth}
        \includegraphics[width=\linewidth]{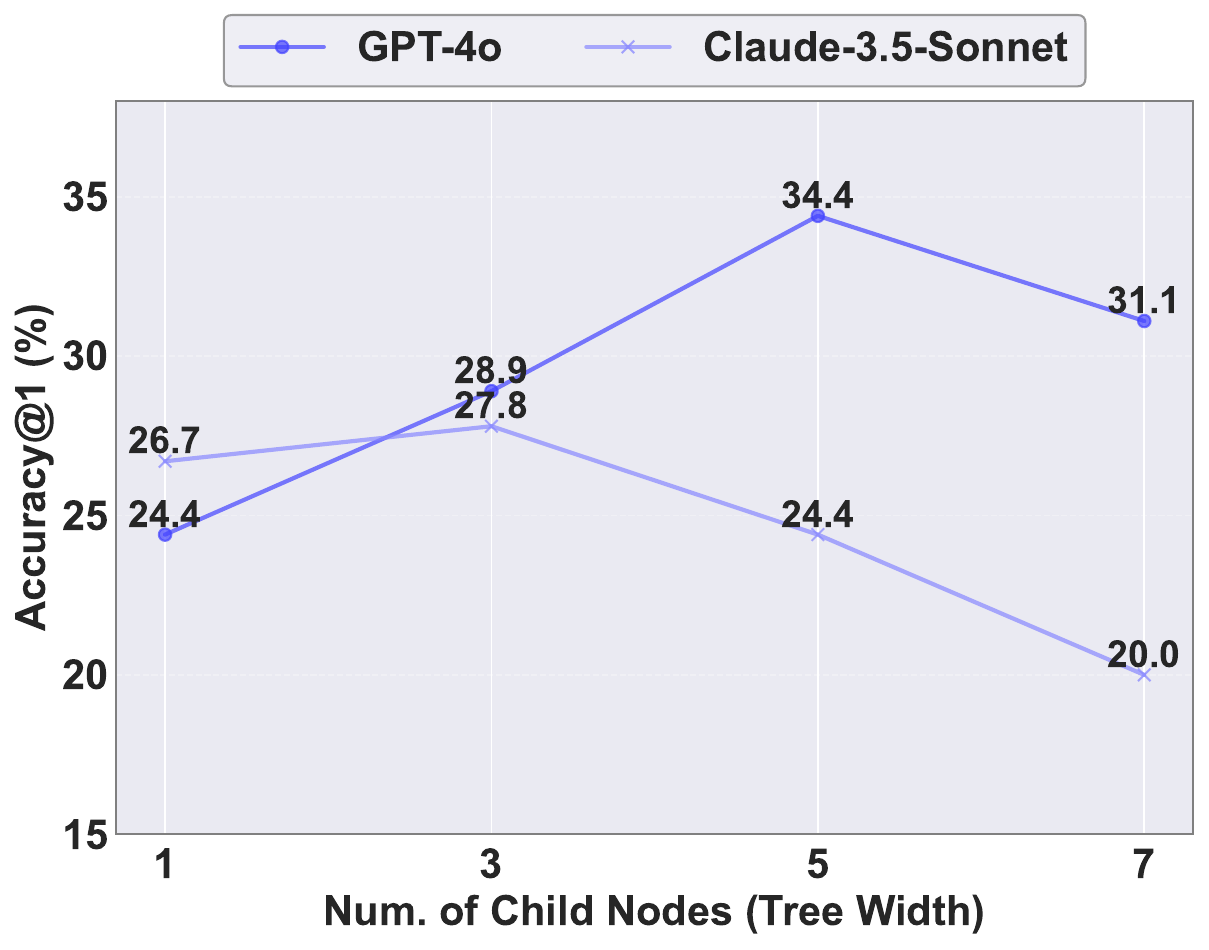}
        \label{figs/tree_width}
    \end{subfigure}
    \vspace{-0.1in}
    \caption{
        (\textbf{\textsc{LEFT}}) The trend of \textit{Pass@1} rate on Visual Navigation as the number of reasoning steps increases. (\textbf{\textsc{right}}) The relationship between the \textit{Accuracy@1} on geometry problems (Geomverse) and tree width for rollout search. \textbf{\textcolor{gray}{We observe that LVLMs significantly benefit from longer reasoning chains, although the effect plateaus rapidly beyond a certain threshold of reasoning steps. The relationship between performance and tree width exhibits a more complex pattern, demonstrating an inverted U-shaped trend with both \textit{GPT-4o} and \textit{Claude-3.5-Sonnet}.}}
    }
    \vspace{-0.15in}
    \label{fig:two_trend}
\end{figure*}

Spatial reasoning, defined as the cognitive capability to interpret spatial object relationships, motion dynamics, and environmental interactions, constitutes a foundational requirement for mission-critical applications such as robotic systems, autonomous navigation, and augmented reality. These domains demand robust integration of visual perception and precise manipulation of spatial-temporal constraints for optimal action planning.

\paragraph{Task Formulation} Building upon the Visualization of Thought (\textit{VoT}) \cite{Wu2024MindsEO} benchmarks, we design two challenging spatial reasoning benchmarks with enhanced complexity as shown in figure \ref{fig:spatial_reasoning}: Visual Navigation and Visual Tiling. We provide detailed materials of the differences between the original VoT benchmark setup and our experimental configuration in Appendix \ref{appendix:spatial_reasoning_task_setting} and additionally provide the mathematical task formulation in appendix \ref{appendix:task_formulation_spatial_reasoning}.

\paragraph{Visual Construction via \textit{\textbf{Executor}}} During task execution, robots deployed in true environments typically receive environmental feedback following each action, which facilitates perception and subsequent decision-making processes. In our methodology, we leverage environmental interaction tools to enhance the model's spatial reasoning capabilities. In each action, we employ an \textit{executor} to implement the corresponding action, and return textual execution feedback and visuospatial hint (\textit{optional}) representing the map state. In the context of (1) Visual Navigation, the visual feedback corresponds to the map including agent's current position; while in (2) Visual Tiling scenarios, it represents the current state of rectangle occupation patterns.

\subsection{Empirical Results}
\label{subsec:empirical_results_spatial_reasoning}

\paragraph{Setup} We evaluate our framework on two spatial reasoning benchmarks: Visual Navigation and Visual Tiling. For Visual Navigation, we create three difficulty levels with increasing map complexity, where the level indicates the $k$ for Visual Navigation as shown in table \ref{tab:visual}. For Visual Tiling, we focus on level-2 (i.e. $k$ = 2) problems with 119 samples. We compare our method against Chain-of-Thought (\textit{CoT}), Visualization of Thought (\textit{VoT}) \cite{Wu2024MindsEO}. As table \ref{tab:visual} indicates, the results from \textit{VoT} with tool interactions (i.e. \textit{Executor}) are also reported, where textual feedbacks are employed but the visual hints are still generated by the model rather from \textit{executor}, consistent with the \textit{VoT} framework. The source of visual hints distinguishes it from our method. We employ the same temperature and \textit{VisuoThink} hyperparameters as section \ref{subsec:empirical_results_goemetry}.

\paragraph{Analysis} 

In spatial reasoning experiments, \textit{VisuoThink} demonstrates significant performance improvements over baseline methods, particularly when augmented with predictive rollout search. As shown in Table \ref{tab:visual}, \textit{VisuoThink} achieves the highest accuracy across all tasks, outperforming both \textit{CoT} and \textit{VoT} baselines. For instance, on the Visual Navigation task, \textit{VisuoThink} on GPT-4o achieves a \textit{93.8}\% accuracy at level-3, compared to \textit{62.5}\% for VoT with an executor and \textit{18.8}\% for CoT. This trend is consistent across different model architectures, including \textit{GPT-4o}, \textit{Qwen2-VL-72B-Instruct}, and \textit{Claude-3.5-sonnet}, highlighting the robustness of our approach. 

Similar to the geometry experiments in Section \ref{sec:geometry}, the integration of tool interactions and multi-step visual reasoning plays a critical role in enhancing performance. The executor's feedback mechanism, which provides visual updates after each action, mirrors the incremental visual refinement seen in geometry tasks, where auxiliary lines are progressively constructed. 
For instance, \textit{VisuoThink} without rollout search demonstrates an average improvement of \textit{34.7\%} on Visual Tiling across diverse models. We observe that while VoT augmented with textual feedback achieves an average increase of \textit{8.1\%}, its performance gain is notably less pronounced compared to \textit{VisuoThink} without rollout search. This underscores the critical role of reliable visual cues in enhancing reasoning capabilities. 
The dynamic interaction allows the model to iteratively refine its reasoning path, leading to more accurate solutions.

%% file: 050-analysis.tex
In this section, we analyze key aspects of \textit{VisuoThink}'s performance. We examine how the length of reasoning chain affects spatial reasoning, the impact of child node expansion in rollout search, and the influence of supervision levels in predictive rollouts across tasks. These insights highlight \textit{VisuoThink}'s effectiveness and suggest future directions for multimodal reasoning frameworks.

\subsection{Could Longer Reasoning Chains Assist LVLMs in Reasoning?}
\label{subsec:analysis:reasoning_steps}

In practical applications of \textit{LVLM}s for spatial reasoning tasks, each tool invocation can be seen as an agent attempting an action in the environment and receiving feedback. Although many attempts may be inaccurate, allowing the model more trial-and-error opportunities before achieving the final goal could potentially enhance its reasoning capabilities. By setting different upper limits on the number of reasoning steps in visual navigation tasks, \textbf{\textcolor{black!80}{we observe a positive correlation between the number of reasoning steps and the model's task completion rate. This suggests that the model indeed benefits from more tool invocations and longer reasoning.}}

However, as the number of reasoning steps increases, the completion rate gradually converges, making further significant improvements challenging. As shown in figure \ref{fig:two_trend} (\textit{left}), for instance, increasing reasoning steps from 10 to 20 resulted in substantial performance gains (\textit{+54.1}\% and \textit{+48.4}\%) across different LVLM architectures (GPT-4o and Claude-3.5-sonnet). However, when reasoning steps were increased from 20 to 40, the performance growth slowed dramatically, dropping to +\textit{6.5}\% and +\textit{2.1}\%, respectively. This phenomenon aligns with expectations, as merely increasing the number of tool invocations does not enable the model to better solve the most challenging samples. This underscores the necessity of techniques like rollout search within the broader context of test scaling.

\subsection{Could Larger Tree Span Enhances \textit{VisuoThink}'s Performance?}
\label{subsec:analysis:child_nodes}

Predictive rollouts enhance the model's reasoning capabilities, which can be viewed as a tangible outcome of successfully expanding the model's reasoning search space. A natural question arises: Can we further improve the model's reasoning performance on benchmarks simply by increasing the number of candidate child nodes at each selection step, i.e., expanding the \textit{tree width}, thereby enhancing model's reasoning capability? To investigate this, we conducted comparative experiments on geometry tasks using GPT-4o and Claude-3.5-sonnet, keeping the depth of the reasoning tree constant while varying the number of candidate child nodes.


As presented in figure \ref{fig:two_trend} (\textit{right}), we observed an inverted U-shaped trend in overall performance as the number of candidate tree nodes increased across different model architectures. Notably, when the number of candidate child nodes equals 1, the model follows a single reasoning path, effectively bypassing predictive rollout search. Contrary to expectations, the performance trend initially rises and then declines. This counterintuitive result can be attributed to the inherent errors in the model's evaluation of child nodes. \textbf{\textcolor{black!80}{Simply and aggressively increasing the tree width leads to confusion in selecting child nodes, which in turn reduces overall reasoning efficiency.}} Thus, an interesting conclusion emerges: we cannot expect to continuously improve model performance by merely increasing the number of child nodes in rollout search.

\begin{figure}[t]
    \centering
    \begin{subfigure}{.45\textwidth}
        \includegraphics[width=\linewidth]{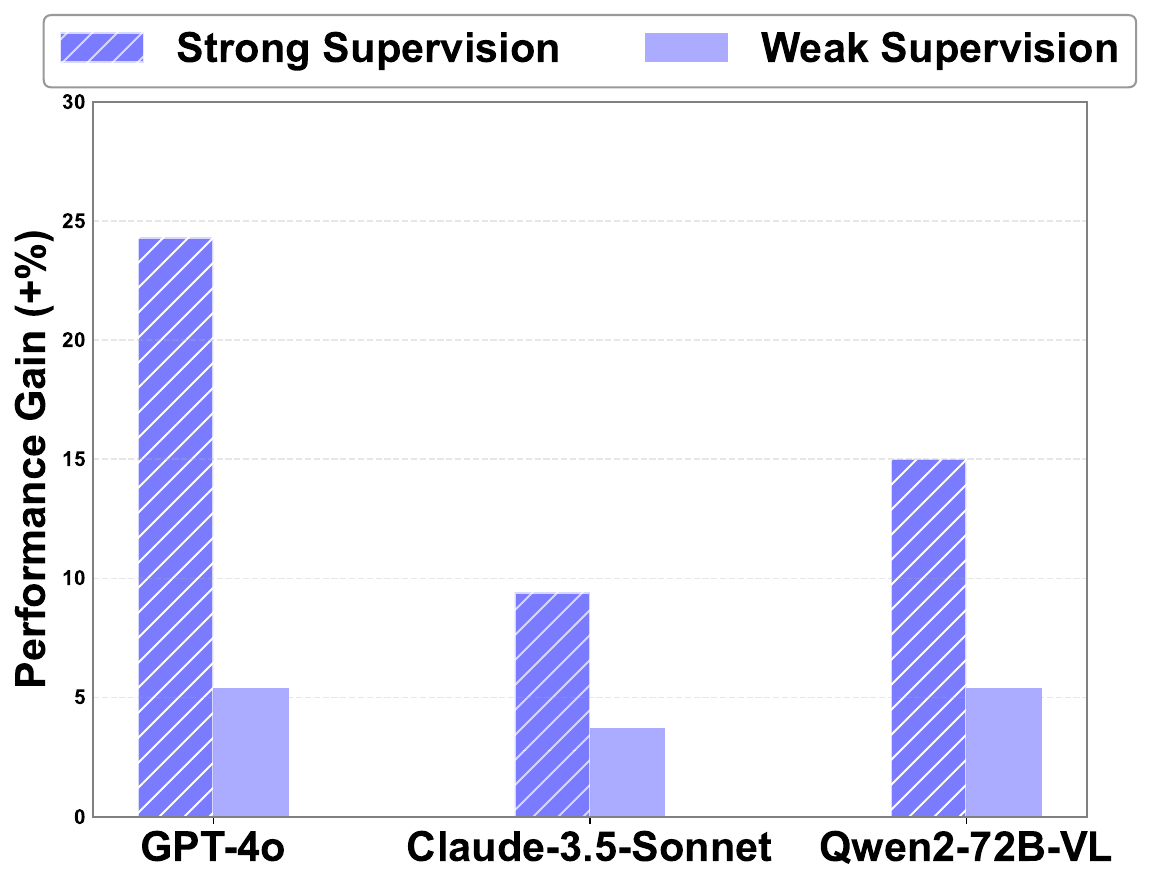}
        \label{figs/api_calls}
    \end{subfigure}
    \vspace{-0.1in}
    \caption{
        The performance gain (\textit{+\%}) on tasks through predictive rollout search. The performance gain is calculated via the performance gap between \textit{VisuoThink (w/o rollout search)} and \textit{VisuoThink}.
    }
    \vspace{-0.1in}
    \label{fig:cluster}
\end{figure}

\subsection{Strong v.s. Weak Supervision in Predictive Rollout Search}
\label{subsec:analysis:supervision}

An intriguing observation is that the strength of guidance provided by predictive rollout results varies between geometry and spatial reasoning tasks. In geometry tasks, the model only receives the final numerical results of the problem, whereas in spatial reasoning tasks, the model has access to visual states of stronger supervision (e.g., \textit{the agent's final position, the position of the destination, etc.}). In other word, predictive rollouts in geometry tasks offer weaker supervision, while those in spatial reasoning tasks provide stronger supervision.

This observation aligns with the findings of the Deepseek R1 report, which highlights that outcome-based supervision in RL can significantly enhance Deepseek-R1-Zero's reasoning capabilities \cite{deepseekai2025deepseekr1}. \textbf{\textcolor{black!80}{The effectiveness of such supervision stems from its strong supervisory signal, and predictive rollouts with strong supervision are more effective in improving model reasoning performance.}} This is further supported by our experimental results, as illustrated in figure \ref{fig:cluster}, where predictive rollouts demonstrated more substantial performance gains in spatial reasoning tasks compared to geometry tasks, across both open-source and closed-source models. The detailed performance gain results are presented in appendix \ref{appendix:performance_gain}.


%% file: appendix.tex
\newpage

\section{Performance Gain of \textbf{\textit{VisuoThink}} Through Predictive Rollout Search}
\label{appendix:performance_gain}

\begin{table*}[t!]
\centering
\resizebox{.8\textwidth}{!}{
    \begin{tabular}{llccc}
    \hline
    \hline
    \textbf{Supervision Type} &   \textbf{Performance Gain}      & \textit{GPT-4o} & \textit{Qwen2-VL-72B} & \textit{Claude-3.5-Sonnet} \\
    \hline
    \multirow{3}{*}{Strong Supervision} &$\Delta$ Visual Navigation (\%) & +16.6 &	+18.9	&+15.5 \\
    &$\Delta$ Visual Tiling (\%) & +31.9	& +11.0 &	+3.3 \\
    & \cellcolor{gray!20} $\Delta$ Average (\%) 
    & \cellcolor{gray!20}+24.3 
    & \cellcolor{gray!20}+15.0 
    & \cellcolor{gray!20}+9.4\\
    \hline
    \multirow{3}{*}{Weak Supervision} &$\Delta$ Geometry3K (\%) & +4.5 &	+6.6 &	+1.1 \\
    &$\Delta$ Geomverse-109 (\%) &+6.2 &	+4.2&	+6.3 \\
    & \cellcolor{gray!20} $\Delta$ Average (\%) 
    & \cellcolor{gray!20}+5.4 
    & \cellcolor{gray!20}+5.4 
    & \cellcolor{gray!20}+3.7\\
    \hline
    \end{tabular}
}
\caption{Detailed performance gain of \textbf{\textit{VisuoThink}} through predictive rollout search on benchmarks from Geometry and Spatial Reasoning over variable \textit{LVLM} models.}
\label{tab:appendix:performace_gain}
\end{table*}

\vspace{0.1in}

This appendix quantifies the performance improvements achieved by integrating predictive rollout search into the \textit{VisuoThink} framework across geometry and spatial reasoning tasks. The performance gain through predictive rollout search is derived by subtracting the performance of \textit{VisuoThink (w/o rollout search)} from those of the \textit{VisuoThink} on models.

\vspace{+0.2in}

As shown in Table \ref{tab:appendix:performace_gain}, tasks with strong supervision (e.g., \textbf{Visual Navigation} and \textbf{Visual Tiling}) exhibit significantly higher gains compared to weak supervision tasks (e.g., \textit{Geometry3K} and \textit{Geomverse-109}). For instance, under strong supervision, Claude-3.5-Sonnet achieves a +\textit{25.1}\% improvement in Visual Navigation, while GPT-4o attains +\textit{16.6}\% in Visual Tiling. In contrast, weak supervision tasks like \textbf{Geomverse-109} only show modest gains (e.g., +\textit{5.4}\% for GPT-4o). 

\vspace{0.2in}

\section{OKSpatial Reasoning Task Setting}
\label{appendix:spatial_reasoning_task_setting}

\begin{table*}[h]
\centering
\resizebox{.8\textwidth}{!}{
    \begin{tabular}{llccc}
    \hline
    \hline
    & \textit{\textbf{Method}}  & Direction & Steps & \textbf{Target} \\
    \hline
    \multirow{2}{*}{ \textbf{Visual Navigation}} &  \textit{VoT} & \Checkmark & \XSolidBrush & Navigate from the starting position \\
    & \cellcolor{gray!20} \textbf{\textit{VisuoThink}} 
    & \cellcolor{gray!20}\Checkmark 
    & \cellcolor{gray!20}\Checkmark 
    & \cellcolor{gray!20} to the destination. \\

    \hline
    \end{tabular}
}
\caption{\textbf{Visual Navigation} task setting differences between \textit{VoT} and \textbf{\textit{VisuoThink}}. }
\label{tab:appendix:spatial_reasoning_task_setting_for_visual_navigation}
\end{table*}

\begin{table*}[!h]
\centering
\resizebox{1.\textwidth}{!}{
    \begin{tabular}{llccccc}
    \hline
    \hline
    & \multirow{2}{*}{\textit{\textbf{Method}}} & \multicolumn{4}{c}{\textbf{Action}} & \multirow{2}{*}{\textbf{Target}} \\
    & & Polyomino Type & Variant Type & Block Positions & Action Type & \\
    \hline
    \multirow{2}{*}{\textbf{Visual Tiling}} & \textit{VoT} & \Checkmark & \Checkmark & \XSolidBrush & \XSolidBrush & \makecell{To identify the correct variant \\ for a polyomino in one action. } \\
    & \cellcolor{gray!20} \textbf{\textit{VisuoThink}} 
    & \cellcolor{gray!20}\Checkmark 
    & \cellcolor{gray!20}\Checkmark 
    & \cellcolor{gray!20}\Checkmark 
    & \cellcolor{gray!20}\Checkmark 
    & \cellcolor{gray!20}\makecell{To fill the rectangle with feasible \\ polyomino variants.} \\

    \hline
    \end{tabular}
}
\caption{\textbf{Visual Tiling} task setting differences between \textit{VoT} and \textbf{\textit{VisuoThink}}.}
\label{tab:appendix:spatial_reasoning_task_setting_for_visual_tiling}
\end{table*}

\vspace{0.1in}

Our formulation extends beyond \textit{VoT}'s basic requirements by mandating LVLMs to generate comprehensive operational specifications - for instance, requiring explicit output of both movement directions and precise step counts at each decision node. This advancement creates more realistic and functionally grounded spatial reasoning evaluations (e.g., \textit{robotic navigation emulation in real world}).

\vspace{+0.2in}

This appendix details the task formulation differences between \textit{VisuoThink} and baseline methods (Table \ref{tab:appendix:spatial_reasoning_task_setting_for_visual_navigation} and Table \ref{tab:appendix:spatial_reasoning_task_setting_for_visual_tiling}). For \textbf{Visual Navigation}, \textit{VisuoThink} requires fine-grained, executable and explicit specification of both direction and step count in action sequences, whereas VoT focuses solely on direction navigation. This formulation mirrors real-world robotic navigation, where precise movement planning is critical. Similarly, in \textbf{Visual Tiling}, \textit{VisuoThink} mandates detailed actions, including polyomino variant types, block positions, and action types (e.g., "fit" or "remove"), while \textit{VoT} simplifies the task by omitting variant specifications.


\section{Task Formulation of Spatial Reasoning Tasks}
\label{appendix:task_formulation_spatial_reasoning}
Building upon \textit{VoT} \cite{Wu2024MindsEO} framework, our challenging benchmarks comprise:

\begin{itemize}
    \item \textbf{Visual Navigation} evaluates LVLMs in a simulated 2D grid environment, where agents must navigate from initial position $\textbf{s}_0$ to destination $\textbf{s}_k$ through obstacle-laden paths. The formal problem is defined by grid map $\mathbf{M}$ containing $k$ interconnected edges $\mathbf{E} = \{\textbf{e}(\textbf{s}_0, \textbf{s}_1), \textbf{e}(\textbf{s}_1, \textbf{s}_2), \dots, \textbf{e}(\textbf{s}_{k-1}, \textbf{s}_k)\} $. The LVLM should generate a sequence of executable actions in \textit{\code{json}} format \( \mathbf{A} = \{(\textbf{d}_0, \textbf{l}_0), (\textbf{d}_1, \textbf{l}_1), \dots, (\textbf{d}_{|\mathbf{A}|-1}, \textbf{l}_{|\mathbf{A}|-1})\} \), where each tuple specifies movement direction \( \textbf{d}_i \) and exact step count \( \textbf{l}_i \), governed by the policy:
   \begin{equation}
   \mathbf{a_t} \sim \mathcal{P} \left(\mathbf{d}_t, \mathbf{l}_t \mid \textbf{A}_{t-1}, \mathbf{M} \right)
   \end{equation}
   \item \textbf{Visual Tiling} is a classic geometric reasoning challenge, this task assesses polyomino composition capabilities within confined rectangular regions $\textbf{R}$ masked by $k$ distinct polyominoes $\mathbf{MP} = \{\textbf{mp}_1, \dots, \textbf{mp}_k\}$. The LVLM must output action sequences $\textbf{a}_t = (\textbf{p}_t, \{\textbf{b}_1, \dots, \textbf{b}_{|B|}\}, \textbf{at}_t)$, where $\textbf{p}_t$ and $\mathbf{B} = \{\textbf{b}_1, \dots, \textbf{b}_{|\mathbf{B}|}\}$ respectively indicate the selected polyomino type and the coordinates of the placement blocks. \( \textbf{at}_t \in \{\textit{\text{fit}}, \textit{\text{remove}}\} \) indicates the action type modifying rectangular state \( \mathbf{R}_t \), thus formalized as: 

   \begin{equation}
       \textbf{a}_t \sim \mathcal{P}\left(\textbf{p}_t, \mathbf{B}, \textbf{at}_t \mid \mathbf{R}_{t-1}, \mathbf{MP}, \textbf{A}_{t-1}\}\right)
   \end{equation} 

   Though the required actions are polyomino variant-aware as shown in table \ref{tab:appendix:spatial_reasoning_task_setting_for_visual_tiling}. As the polyomino variant type is implicitly expressed in the block positions, LVLM does not need to explicitly output it in actions anymore.
\end{itemize}

\vspace{1.8in}

\section{Model and \textbf{\textit{VisuoThink}} Hyperparameters}
\label{appendix:hyperparameter}

We detail the model and \textit{VisuoThink} Hyperparameters:


\paragraph{Model Hyperparameters} To ensure experimental fairness, we uniformly constrained the number of reasoning steps (i.e., $\tau$, \textit{the depth of the reasoning tree}) to \textit{10} across all experiments. During predictive rollout search, we set the number of sampled child nodes to \textit{3}, and we discuss its impact in section \ref{subsec:analysis:child_nodes}.


\paragraph{\textit{VisuoThink} Hyperparameters} While \textit{VisuoThink} employed a temperature of \textit{0.8} when sampling child nodes, all other model invocations, including the baselines (e.g. \textit{CoT}, \textit{VoT}, \textit{VisualSketchpad}, \textit{VisuoThink} w/o rollout search), were conducted with temperature set to \textit{0} for frontier performance. During the voting phase, we similarly maintained a temperature of \textit{0} and implemented single-vote sampling, which not only reduced computational overhead in terms of model calls but also achieved comparable performance.

\section{Geomverse-109 Problem Generation Trajectory}
\label{appendix:geomverse}

We establish a pipeline translating textual problems into problems with matplotlib-executable code. Beyond the \textbf{Geometry3K} \cite{lu2021intergpsinterpretablegeometryproblem} dataset (\textit{48 problems}) utilized in Sketchpad, we incorporate the D2 subset of Geomverse \cite{geomverse} to construct an slightly bigger dataset \textbf{Geomverse-109} (\textit{90 problems}). The original Geomverse dataset crucially includes annotated point coordinates essential for systematic problem synthesis. During the data synthesis phase, we first randomly choose 109 problems, then LVLMs generate corresponding high-quality Python code through LLM self-reflection \cite{reflexion}, then we filter out problems with poor diagram quality.

%% file: acl_latex.bbl
\begin{thebibliography}{38}
\providecommand{\natexlab}[1]{#1}

\bibitem[{Amini et~al.(2024)Amini, Vieira, and Cotterell}]{Amini2024VariationalBA}
Afra Amini, Tim Vieira, and Ryan Cotterell. 2024.
\newblock \href {https://api.semanticscholar.org/CorpusID:271051300} {Variational best-of-n alignment}.
\newblock \emph{ArXiv}, abs/2407.06057.

\bibitem[{Chen et~al.(2024)Chen, Liao, Li, and Fan}]{Chen2024AlphaMathAZ}
Guoxin Chen, Minpeng Liao, Chengxi Li, and Kai Fan. 2024.
\newblock \href {https://api.semanticscholar.org/CorpusID:269605484} {Alphamath almost zero: process supervision without process}.
\newblock \emph{ArXiv}, abs/2405.03553.

\bibitem[{Cherian et~al.(2024)Cherian, Peng, Lohit, Matthiesen, Smith, and Tenenbaum}]{cherian2024evaluating}
Anoop Cherian, Kuan-Chuan Peng, Suhas Lohit, Joanna Matthiesen, Kevin Smith, and Joshua~B Tenenbaum. 2024.
\newblock Evaluating large vision-and-language models on children's mathematical olympiads.
\newblock \emph{arXiv preprint arXiv:2406.15736}.

\bibitem[{DeepSeek-AI(2025)}]{deepseekai2025deepseekr1}
DeepSeek-AI. 2025.
\newblock \href {https://arxiv.org/abs/2501.12948} {Deepseek-r1: Incentivizing reasoning capability in llms via reinforcement learning}.
\newblock \emph{Preprint}, arXiv:2501.12948.

\bibitem[{Du et~al.(2025)Du, Liu, Li, Zhao, Huo, Wang, Chen, Liu, Wang, and Wen}]{Du2025VirgoAP}
Yifan Du, Zikang Liu, Yifan Li, Wayne~Xin Zhao, Yuqi Huo, Bingning Wang, Weipeng Chen, Zheng Liu, Zhongyuan Wang, and Jiahui Wen. 2025.
\newblock \href {https://api.semanticscholar.org/CorpusID:275323902} {Virgo: A preliminary exploration on reproducing o1-like mllm}.

\bibitem[{Feng et~al.(2023)Feng, Wan, Wen, Wen, Zhang, and Wang}]{Feng2023AlphazerolikeTC}
Xidong Feng, Ziyu Wan, Muning Wen, Ying Wen, Weinan Zhang, and Jun Wang. 2023.
\newblock \href {https://api.semanticscholar.org/CorpusID:263310590} {Alphazero-like tree-search can guide large language model decoding and training}.
\newblock \emph{ArXiv}, abs/2309.17179.

\bibitem[{Gao et~al.(2024)Gao, Chen, Zhang, Fu, Shen, Zhang, Zhang, Zheng, Sun, Cao, and Ji}]{Gao2024CantorIM}
Timin Gao, Peixian Chen, Mengdan Zhang, Chaoyou Fu, Yunhang Shen, Yan Zhang, Shengchuan Zhang, Xiawu Zheng, Xing Sun, Liujuan Cao, and Rongrong Ji. 2024.
\newblock \href {https://api.semanticscholar.org/CorpusID:269362481} {Cantor: Inspiring multimodal chain-of-thought of mllm}.
\newblock \emph{ArXiv}, abs/2404.16033.

\bibitem[{Gui et~al.(2024)Gui, Garbacea, and Veitch}]{Gui2024BoNBoNAF}
Lin Gui, Cristina Garbacea, and Victor Veitch. 2024.
\newblock \href {https://api.semanticscholar.org/CorpusID:270213066} {Bonbon alignment for large language models and the sweetness of best-of-n sampling}.
\newblock \emph{ArXiv}, abs/2406.00832.

\bibitem[{Hu et~al.(2024)Hu, Shi, Fu, Roth, Ostendorf, Zettlemoyer, Smith, and Krishna}]{Hu2024VisualSS}
Yushi Hu, Weijia Shi, Xingyu Fu, Dan Roth, Mari Ostendorf, Luke~S. Zettlemoyer, Noah~A. Smith, and Ranjay Krishna. 2024.
\newblock \href {https://api.semanticscholar.org/CorpusID:270440440} {Visual sketchpad: Sketching as a visual chain of thought for multimodal language models}.
\newblock \emph{ArXiv}, abs/2406.09403.

\bibitem[{Kahneman(2011)}]{kahneman2011thinking}
Daniel Kahneman. 2011.
\newblock Thinking, fast and slow.
\newblock \emph{Farrar, Straus and Giroux}.

\bibitem[{Kazemi et~al.(2023)Kazemi, Alvari, Anand, Wu, Chen, and Soricut}]{geomverse}
Mehran Kazemi, Hamidreza Alvari, Ankit Anand, Jialin Wu, Xi~Chen, and Radu Soricut. 2023.
\newblock \href {https://arxiv.org/abs/2312.12241} {Geomverse: A systematic evaluation of large models for geometric reasoning}.
\newblock \emph{Preprint}, arXiv:2312.12241.

\bibitem[{Lei et~al.(2024)Lei, Yang, Chen, Li, and Liu}]{Lei2024ScaffoldingCT}
Xuanyu Lei, Zonghan Yang, Xinrui Chen, Peng Li, and Yang Liu. 2024.
\newblock \href {https://api.semanticscholar.org/CorpusID:267750933} {Scaffolding coordinates to promote vision-language coordination in large multi-modal models}.
\newblock In \emph{International Conference on Computational Linguistics}.

\bibitem[{Li et~al.(2025)Li, Wu, Zhang, Xia, Mao, Dong, Vuli'c, and Wei}]{Li2025ImagineWR}
Chengzu Li, Wenshan Wu, Huanyu Zhang, Yan Xia, Shaoguang Mao, Li~Dong, Ivan Vuli'c, and Furu Wei. 2025.
\newblock \href {https://api.semanticscholar.org/CorpusID:275471612} {Imagine while reasoning in space: Multimodal visualization-of-thought}.

\bibitem[{Liu et~al.(2023)Liu, Cohen, Pasunuru, Choi, Hajishirzi, and Celikyilmaz}]{Liu2023DontTA}
Jiacheng Liu, Andrew Cohen, Ramakanth Pasunuru, Yejin Choi, Hannaneh Hajishirzi, and Asli Celikyilmaz. 2023.
\newblock \href {https://api.semanticscholar.org/CorpusID:262824527} {Don't throw away your value model! generating more preferable text with value-guided monte-carlo tree search decoding}.

\bibitem[{Lu et~al.(2021)Lu, Gong, Jiang, Qiu, Huang, Liang, and Zhu}]{lu2021intergpsinterpretablegeometryproblem}
Pan Lu, Ran Gong, Shibiao Jiang, Liang Qiu, Siyuan Huang, Xiaodan Liang, and Song-Chun Zhu. 2021.
\newblock \href {https://arxiv.org/abs/2105.04165} {Inter-gps: Interpretable geometry problem solving with formal language and symbolic reasoning}.
\newblock \emph{Preprint}, arXiv:2105.04165.

\bibitem[{Mitra et~al.(2023)Mitra, Huang, Darrell, and Herzig}]{Mitra2023CompositionalCP}
Chancharik Mitra, Brandon Huang, Trevor Darrell, and Roei Herzig. 2023.
\newblock \href {https://api.semanticscholar.org/CorpusID:265498786} {Compositional chain-of-thought prompting for large multimodal models}.
\newblock \emph{2024 IEEE/CVF Conference on Computer Vision and Pattern Recognition (CVPR)}, pages 14420--14431.

\bibitem[{Mondal et~al.(2024)Mondal, Modi, Panda, Singh, and Rao}]{Mondal2024KAMCoTKA}
Debjyoti Mondal, Suraj Modi, Subhadarshi Panda, Rituraj Singh, and Godawari~Sudhakar Rao. 2024.
\newblock \href {https://api.semanticscholar.org/CorpusID:267095090} {Kam-cot: Knowledge augmented multimodal chain-of-thoughts reasoning}.
\newblock In \emph{AAAI Conference on Artificial Intelligence}.

\bibitem[{OpenAI(2024{\natexlab{a}})}]{openai2024gpt4ocard}
OpenAI. 2024{\natexlab{a}}.
\newblock \href {https://arxiv.org/abs/2410.21276} {Gpt-4o system card}.
\newblock \emph{Preprint}, arXiv:2410.21276.

\bibitem[{OpenAI(2024{\natexlab{b}})}]{openai2024o1}
OpenAI. 2024{\natexlab{b}}.
\newblock \href {https://openai.com/index/learning-to-reason-with-llms/} {Learning to reason with llms}.

\bibitem[{Qiao et~al.(2024)Qiao, Tan, Dong, Wu, Sun, Song, GongQue, Lei, Wei, Zhang et~al.}]{qiao2024we}
Runqi Qiao, Qiuna Tan, Guanting Dong, Minhui Wu, Chong Sun, Xiaoshuai Song, Zhuoma GongQue, Shanglin Lei, Zhe Wei, Miaoxuan Zhang, and 1 others. 2024.
\newblock We-math: Does your large multimodal model achieve human-like mathematical reasoning?
\newblock \emph{arXiv preprint arXiv:2407.01284}.

\bibitem[{Ramakrishnan et~al.(2024)Ramakrishnan, Wijmans, Kraehenbuehl, and Koltun}]{ramakrishnan2024doesspatialcognitionemerge}
Santhosh~Kumar Ramakrishnan, Erik Wijmans, Philipp Kraehenbuehl, and Vladlen Koltun. 2024.
\newblock \href {https://arxiv.org/abs/2410.06468} {Does spatial cognition emerge in frontier models?}
\newblock \emph{Preprint}, arXiv:2410.06468.

\bibitem[{Shinn et~al.(2023)Shinn, Cassano, Berman, Gopinath, Narasimhan, and Yao}]{reflexion}
Noah Shinn, Federico Cassano, Edward Berman, Ashwin Gopinath, Karthik Narasimhan, and Shunyu Yao. 2023.
\newblock \href {https://arxiv.org/abs/2303.11366} {Reflexion: Language agents with verbal reinforcement learning}.
\newblock \emph{Preprint}, arXiv:2303.11366.

\bibitem[{Snell et~al.(2024)Snell, Lee, Xu, and Kumar}]{Snell2024ScalingLT}
Charlie Snell, Jaehoon Lee, Kelvin Xu, and Aviral Kumar. 2024.
\newblock \href {https://api.semanticscholar.org/CorpusID:271719990} {Scaling llm test-time compute optimally can be more effective than scaling model parameters}.
\newblock \emph{ArXiv}, abs/2408.03314.

\bibitem[{Sun et~al.(2024)Sun, Haider, Zhang, Yang, Qiu, Yin, Wang, Bartlett, and Zanette}]{Sun2024FastBD}
Hanshi Sun, Momin Haider, Ruiqi Zhang, Huitao Yang, Jiahao Qiu, Ming Yin, Mengdi Wang, Peter Bartlett, and Andrea Zanette. 2024.
\newblock \href {https://api.semanticscholar.org/CorpusID:273654642} {Fast best-of-n decoding via speculative rejection}.
\newblock \emph{ArXiv}, abs/2410.20290.

\bibitem[{Team(2024)}]{geminiteam2024gemini15}
Gemini Team. 2024.
\newblock \href {https://arxiv.org/abs/2403.05530} {Gemini 1.5: Unlocking multimodal understanding across millions of tokens of context}.
\newblock \emph{Preprint}, arXiv:2403.05530.

\bibitem[{Thawakar et~al.(2025)Thawakar, Dissanayake, More, Thawkar, Heakl, Ahsan, Li, Zumri, Lahoud, Anwer, Cholakkal, Laptev, Shah, Khan, and Khan}]{Thawakar2025LlamaVo1RS}
Omkar Thawakar, Dinura Dissanayake, Ketan More, Ritesh Thawkar, Ahmed Heakl, Noor Ahsan, Yuhao Li, Mohammed Zumri, Jean Lahoud, Rao~Muhammad Anwer, Hisham Cholakkal, Ivan Laptev, Mubarak Shah, Fahad~Shahbaz Khan, and Salman~H. Khan. 2025.
\newblock \href {https://api.semanticscholar.org/CorpusID:275458766} {Llamav-o1: Rethinking step-by-step visual reasoning in llms}.

\bibitem[{Wu et~al.(2024)Wu, Mao, Zhang, Xia, Dong, Cui, and Wei}]{Wu2024MindsEO}
Wenshan Wu, Shaoguang Mao, Yadong Zhang, Yan Xia, Li~Dong, Lei Cui, and Furu Wei. 2024.
\newblock \href {https://api.semanticscholar.org/CorpusID:268889526} {Mind's eye of llms: Visualization-of-thought elicits spatial reasoning in large language models}.

\bibitem[{Xiang et~al.(2024)Xiang, Liu, Jiang, Nie, Huang, Fan, Li, Huang, Zeng, Han, Hong, Xu, and Liang}]{xiang2024atomthinkslowthinkingframework}
Kun Xiang, Zhili Liu, Zihao Jiang, Yunshuang Nie, Runhui Huang, Haoxiang Fan, Hanhui Li, Weiran Huang, Yihan Zeng, Jianhua Han, Lanqing Hong, Hang Xu, and Xiaodan Liang. 2024.
\newblock \href {https://arxiv.org/abs/2411.11930} {Atomthink: A slow thinking framework for multimodal mathematical reasoning}.
\newblock \emph{Preprint}, arXiv:2411.11930.

\bibitem[{Xie et~al.(2023)Xie, Kawaguchi, Zhao, Zhao, Kan, He, and Xie}]{Xie2023SelfEvaluationGB}
Yuxi Xie, Kenji Kawaguchi, Yiran Zhao, Xu~Zhao, MingSung Kan, Junxian He, and Qizhe Xie. 2023.
\newblock \href {https://api.semanticscholar.org/CorpusID:258426922} {Self-evaluation guided beam search for reasoning}.
\newblock In \emph{Neural Information Processing Systems}.

\bibitem[{Xu et~al.(2024)Xu, Jin, Li, Song, Sun, and Yuan}]{Xu2024LLaVACoTLV}
Guowei Xu, Peng Jin, Hao Li, Yibing Song, Lichao Sun, and Li~Yuan. 2024.
\newblock \href {https://api.semanticscholar.org/CorpusID:274116688} {Llava-cot: Let vision language models reason step-by-step}.
\newblock \emph{ArXiv}, abs/2411.10440.

\bibitem[{Yang et~al.(2023)Yang, Li, Wang, Lin, Azarnasab, Ahmed, Liu, Liu, Zeng, and Wang}]{Yang2023MMREACTPC}
Zhengyuan Yang, Linjie Li, Jianfeng Wang, Kevin Lin, Ehsan Azarnasab, Faisal Ahmed, Zicheng Liu, Ce~Liu, Michael Zeng, and Lijuan Wang. 2023.
\newblock \href {https://api.semanticscholar.org/CorpusID:257637012} {Mm-react: Prompting chatgpt for multimodal reasoning and action}.
\newblock \emph{ArXiv}, abs/2303.11381.

\bibitem[{Yao et~al.(2024)Yao, Huang, Wu, Zhang, Wang, Liu, Wang, Song, Feng, Shen, and Tao}]{Yao2024MulberryEM}
Huanjin Yao, Jiaxing Huang, Wenhao Wu, Jingyi Zhang, Yibo Wang, Shunyu Liu, Yingjie Wang, Yuxin Song, Haocheng Feng, Li~Shen, and Dacheng Tao. 2024.
\newblock \href {https://api.semanticscholar.org/CorpusID:274992111} {Mulberry: Empowering mllm with o1-like reasoning and reflection via collective monte carlo tree search}.
\newblock \emph{ArXiv}, abs/2412.18319.

\bibitem[{Yao et~al.(2023)Yao, Zhao, Yu, Du, Shafran, Narasimhan, and Cao}]{react}
Shunyu Yao, Jeffrey Zhao, Dian Yu, Nan Du, Izhak Shafran, Karthik Narasimhan, and Yuan Cao. 2023.
\newblock \href {https://arxiv.org/abs/2210.03629} {React: Synergizing reasoning and acting in language models}.
\newblock \emph{Preprint}, arXiv:2210.03629.

\bibitem[{Yu et~al.(2023)Yu, Gao, and Wang}]{Yu2023OVMOV}
Fei Yu, Anningzhe Gao, and Benyou Wang. 2023.
\newblock \href {https://api.semanticscholar.org/CorpusID:265221057} {Ovm, outcome-supervised value models for planning in mathematical reasoning}.
\newblock In \emph{NAACL-HLT}.

\bibitem[{Zeng et~al.(2024)Zeng, Cheng, Yin, Wang, Li, Zhou, Guo, Huang, and Qiu}]{zeng2024scalingsearchlearningroadmap}
Zhiyuan Zeng, Qinyuan Cheng, Zhangyue Yin, Bo~Wang, Shimin Li, Yunhua Zhou, Qipeng Guo, Xuanjing Huang, and Xipeng Qiu. 2024.
\newblock \href {https://arxiv.org/abs/2412.14135} {Scaling of search and learning: A roadmap to reproduce o1 from reinforcement learning perspective}.
\newblock \emph{Preprint}, arXiv:2412.14135.

\bibitem[{Zhang et~al.(2023)Zhang, Zhang, Li, Zhao, Karypis, and Smola}]{Zhang2023MultimodalCR}
Zhuosheng Zhang, Aston Zhang, Mu~Li, Hai Zhao, George Karypis, and Alexander~J. Smola. 2023.
\newblock \href {https://api.semanticscholar.org/CorpusID:256504063} {Multimodal chain-of-thought reasoning in language models}.
\newblock \emph{Trans. Mach. Learn. Res.}, 2024.

\bibitem[{Zhao et~al.(2023)Zhao, Rao, Liu, Liu, Zhou, and Lu}]{zhao2023unleashingtexttoimagediffusionmodels}
Wenliang Zhao, Yongming Rao, Zuyan Liu, Benlin Liu, Jie Zhou, and Jiwen Lu. 2023.
\newblock \href {https://arxiv.org/abs/2303.02153} {Unleashing text-to-image diffusion models for visual perception}.
\newblock \emph{Preprint}, arXiv:2303.02153.

\bibitem[{Zhou et~al.(2024)Zhou, Zhou, Hu, Lu, Gao, and Zhang}]{Zhou2024ImageofThoughtPF}
Qiji Zhou, Ruochen Zhou, Zike Hu, Panzhong Lu, Siyang Gao, and Yue Zhang. 2024.
\newblock \href {https://api.semanticscholar.org/CorpusID:269982092} {Image-of-thought prompting for visual reasoning refinement in multimodal large language models}.
\newblock \emph{ArXiv}, abs/2405.13872.

\end{thebibliography}
